\documentclass[10pt,twocolumn,letterpaper]{article}

\usepackage{iccv}
\usepackage{times}
\usepackage{epsfig}
\usepackage{graphicx}
\usepackage{amsmath}
\usepackage{amssymb}
\usepackage{subfigure}

\DeclareMathOperator*{\sign}{sign}

\newcommand{\ft}{f_{\theta}}
\newcommand{\ts}{\theta^{\star}}

\RequirePackage{amsmath}
\RequirePackage{amssymb}
\ifx\proof\undefined 
\RequirePackage{amsthm}
\fi
\RequirePackage{bm} 
\RequirePackage{url}

\let\hat\widehat

\ifx\BlackBox\undefined
\newcommand{\BlackBox}{\rule{1.5ex}{1.5ex}}  
\fi

\ifx\QED\undefined
\def\QED{~\rule[-1pt]{5pt}{5pt}\par\medskip}
\fi

\ifx\proof\undefined
\newenvironment{proof}{\par\noindent{\bf Proof\ }}{\hfill\BlackBox\\[2mm]}
\fi

\ifx\theorem\undefined
\newtheorem{theorem}{Theorem}
\fi
\ifx\assumption\undefined
\newtheorem{assumption}[theorem]{Assumption}
\fi
\ifx\condition\undefined
\newtheorem{condition}{Condition}
\fi
\ifx\example\undefined
\newtheorem{example}{Example}
\fi
\ifx\property\undefined
\newtheorem{property}{Property}
\fi
\ifx\lemma\undefined
\newtheorem{lemma}[theorem]{Lemma}
\fi
\ifx\proposition\undefined
\newtheorem{proposition}[theorem]{Proposition}
\fi
\ifx\remark\undefined
\newtheorem{remark}{Remark}
\fi
\ifx\corollary\undefined
\newtheorem{corollary}[theorem]{Corollary}
\fi
\ifx\definition\undefined
\newtheorem{definition}[theorem]{Definition}
\fi
\ifx\conjecture\undefined
\newtheorem{conjecture}[theorem]{Conjecture}
\fi
\ifx\fact\undefined
\newtheorem{fact}[theorem]{Fact}
\fi
\ifx\claim\undefined
\newtheorem{claim}[theorem]{Claim}
\fi

\makeatletter
\newtheorem*{rep@theorem}{\rep@title}
\newcommand{\newreptheorem}[2]{%
\newenvironment{rep#1}[1]{%
 \def\rep@title{#2 \ref{##1}}%
 \begin{rep@theorem}}%
 {\end{rep@theorem}}}
\makeatother
\newreptheorem{theorem}{Theorem}
\newreptheorem{lemma}{Lemma}
\newreptheorem{corollary}{Corollary}
\newreptheorem{proposition}{Proposition}

\usepackage[pagebackref=true,breaklinks=true,letterpaper=true,colorlinks,bookmarks=false]{hyperref}

\iccvfinalcopy 

\def\iccvPaperID{11465} 

\ificcvfinal\fi

\begin{document}

\title{Deep Nonparametric Convexified Filtering for Computational Photography, Image Synthesis and Adversarial Defense}

\author{Jianqiao Wangni\\
Zhipu AI\\
{\tt\small zjnqha@gmail.com}
}

\maketitle
\ificcvfinal\thispagestyle{empty}\fi
\begin{abstract}
We aim to provide a general framework of for computational photography that recovers the real scene from imperfect images, via the Deep Nonparametric Convexified Filtering (DNCF). It is consists of a  nonparametric deep network to resemble the physical equations behind the image formation, such as denoising, super-resolution, inpainting, and flash.  DNCF has no parameterization dependent on training data, therefore has a strong generalization and robustness to adversarial image manipulation.  During inference, we also encourage the network parameters to be nonnegative and create a bi-convex function on the input and parameters, and this adapts to second-order optimization algorithms with insufficient running time, having 10X acceleration over DIP. 
With these tools, we empirically verify its capability to defend image classification deep networks against adversary attack algorithms in real-time. 
\end{abstract}

\newcommand{\obj}{\mathcal L}
\newcommand{\gt}{\partial \mathcal L/\partial {y^t}}
\newcommand{\ys}{y^{\star}}
\newcommand{\dl}{\nabla \mathcal L}
\newcommand{\pro}{\textit{Proj}_{\Phi}}

\section{Introduction}
Computational photography aims to recover real scenes from imperfect images captured by cameras. Understanding the physics behind the image formation, like gain control, aperture, exposure time and depth of focus, etc, are fundamental works in the area. Besides the physics principles that are clear and universal, lots of factors like object depth, are treated as a random variable. For computational photography, we usually solve an inverse problem with a statistical prior of these factors. The characterization of the prior is thus an important task and contributes significantly to the selection of a suitable algorithm. Take image denoising, for example, the Wiener filter works better for Gaussian white noise, and the median filter fits the salt-and-pepper noise \cite{chan2005image}. Besides, any estimation or pre-calibration of the noise level function \cite{mccamy1976color} \cite{grossberg2004modeling}  helps with denoising. Meanwhile, some other imaging factors are more difficult to estimate from limited images. Take the single image blurring task as an example; there is no proper prior knowledge of the essential variables like object depth, which directly affects the point spread function (PSF) of out-of-focus blur kernel; object and camera movement are also unknown. There were some works in the line of deep learning that learn a statistical prior from massive training data for photography tasks, e.g, super-resolution \cite{dong2015image} , image dehazing \cite{cai2016dehazenet}, deblurring \cite{tao2018scale} and denoising \cite{zhang2017learning}\cite{xie2012image}. However, rigorously, such approaches have to rely on the assumption that testing images resemble training images to ensure generalization.

\begin{figure*}[htbp]
	\centering
	\subfigure 
	{\includegraphics[width=0.9\textwidth]{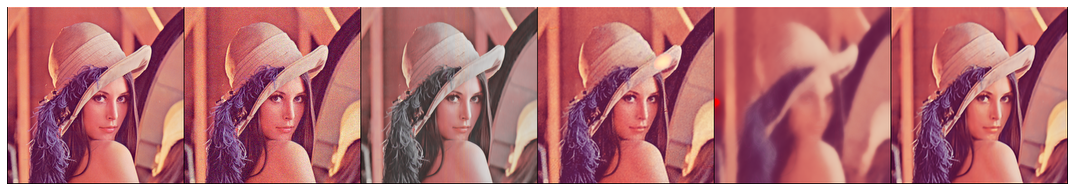}}
	\subfigure 
	{\includegraphics[width=0.9\textwidth]{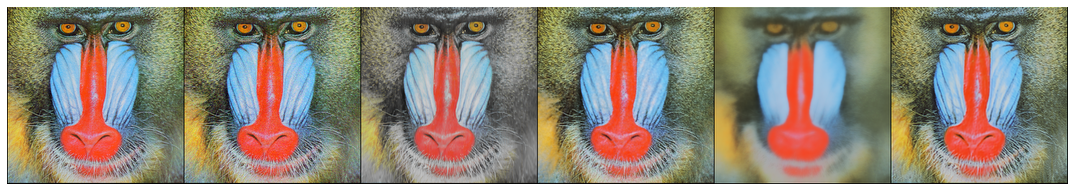}}
		\subfigure 
	{\includegraphics[width=0.9\textwidth]{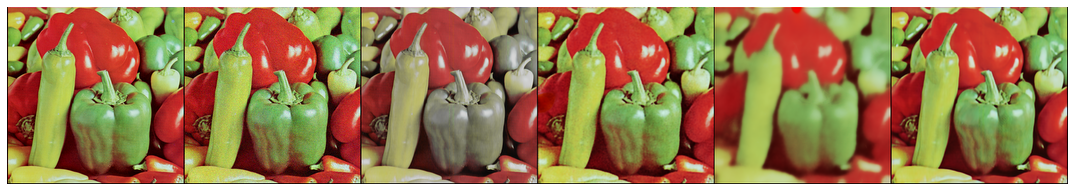}}
	\subfigure 
	{\includegraphics[width=0.9\textwidth]{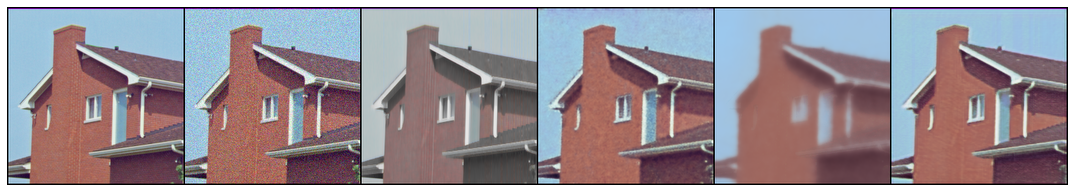}}
	\caption{ Image denoising. Left to right: clean, noisy, TV \cite{chambolle2004algorithm}, DIP (92.7 sec), DeepRED\cite{Mataev_2019_ICCV} +TV (68.1 sec), DNCF (11.9 sec). 
	}
	\label{fig:denoise}
	\vspace{-10pt}
\end{figure*}
We view the aforementioned computational photography algorithms  as \textit{nonparametric}, that their parameters are specific to each image, and do not depend on training data. On a very different track of applications, where the images are used in semantic tasks like classification with deep neural networks, the researchers found the networks are extremely easy to be fooled, and the images seem to be deceptive in the eyes of deep learning \cite{biggio2012poisoning} \cite{szegedy2013intriguing}\cite{goodfellow2014explaining} \cite{kurakin2016adversarial}\cite{carlini2017towards}. Without precaution, an adversary party can manually create images to make the deep network having incorrect predictions with high-confidence, even those images were only manipulated in details that human eyes are unable to notice. We understand this from the intrinsic nature of the deep networks, which are mostly over-parameterized, 
and most math operations are differentiable. It is easier to get better performance for deep learning, as these parameters are practical to optimize. This can be a double-edged sword, since if one party can easily train the parameters of the network, then the adversary party can easily manipulate the images to the wrong side. This inspires us to think that any deep network that has no specific parameters will leave no chances of attack to the adversary, by being exactly the opposite of conventional deep learning approaches. 

A recent method named Deep Image Prior (DIP)\cite{ulyanov2018deep}\cite{cheng2019bayesian}\cite{Mataev_2019_ICCV}\cite{gadelha2019shape} addresses the problem by describing an image prior implicitly by the network structure, and it no longer needs any training. DIP assumes that each image has an underlying parameterization, which is specific to the individual image itself. The target image $I_t$ is synthesized by a neural network $f$ with $\theta$ as its parameters, and it is generated from a random vector variable $z$.
\begin{align}
    I_t = f_{\ts}(z).
\end{align}
The network may consist of convolution, downsampling, deconvolution, and upsampling layers specifically for different tasks. 
Simply defining the structure of $f$, the target images $I_t$ is reconstructed through gradient-based optimization w.r.t. $\theta$, based on the observed source image $I_s$,
\begin{align}\label{eq1}
    \ts=\arg\min_{\theta} \mathcal L \left( I_s, f_{\theta}(z)\right).
\end{align}
where the loss function $\mathcal L$ can be square $\ell_2$ distance or total variation. The interpretation behind the formulation resembles that the target image should be an optimal point of a regularized energy function $E(I;I_s)+ \mathcal S(I)$, 
the regularization $\mathcal S(I)=0$ for $\exists z, I=f_{\theta}(z)$ and $\mathcal S(I)= +\infty$ otherwise, which states that $I$ is constrained by the network structure, and the prior is defined by its expressiveness.

We study a better network structure of $f$ that better fits the physical process and a faster inference pipeline. Start with considering the simplicity of the loss function driven by physics, e.g.; we observe a deblurring model that adopts a denoising feature as
\begin{align}\label{eq:blur}
 \hat I_t,\hat{ \mathcal K} =   \min_{I_t, \mathcal K}\| I_t \ast  \mathcal K -I_s \|^2,\quad 
\end{align}
where $\mathcal K$ is the unknown point spread function (PSF). This blind deconvolution model is though nonconvex w.r.t the unknown variables $\{I_t,\mathcal K\}$, once the PSF is known, the loss function is convex w.r.t the target image. The neural networks are typically composed of many layers, being inherently highly nonlinear and nonconvex w.r.t to the network parameter and inputs, and may consist of up to millions of parameters, so they lose the simplicity of physical equations, many of which are linear or convex. In this paper, we try to explore a framework that are as expressive as deep networks and also being explainable and nonparametric as blind deconvolution.

\begin{figure*}[htbp]
	\centering
	\subfigure 
	{\includegraphics[width=0.9\textwidth]{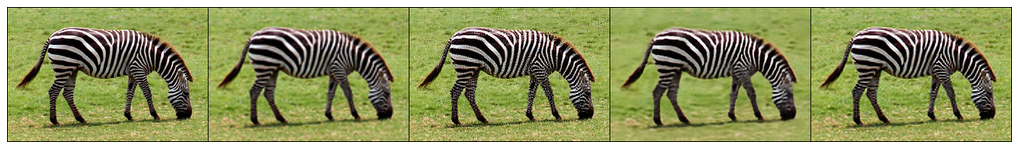}}
	\subfigure 
{\includegraphics[width=0.9\textwidth]{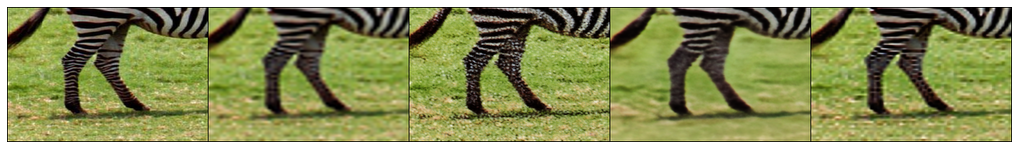}}
	\subfigure 
	{\includegraphics[width=0.9\textwidth]{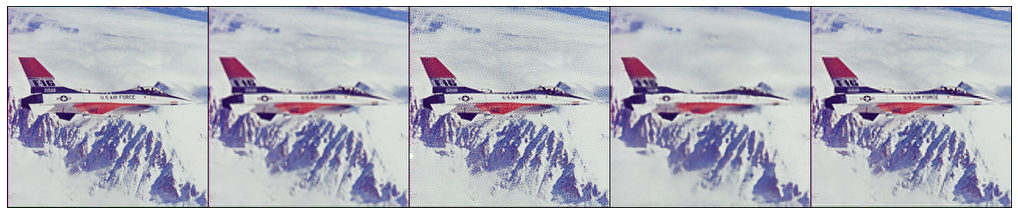}}
	\subfigure 
{\includegraphics[width=0.9\textwidth]{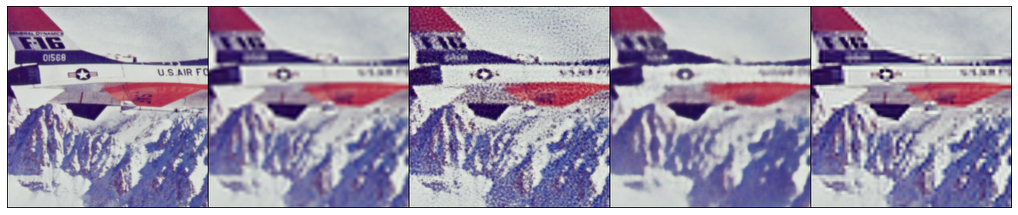}}
	\caption{ Visualization of Super-resolution with limited running time.  Left to right: high resolution, low resolution,   DIP (353 sec), DeepRED + TV (112.1 sec), DNCF (12.0 sec). 
	}
	\label{fig:SR}
\end{figure*}

\section{Approach}
This section presents our nonparametric deep network approach to solve computational photography. 
Our first consideration is to have a decomposed neural network of two meaningful parts: a physical interference model $f$ sequentially connects to the generative model $g$, thus $I_s= f_{\theta}(g_{\theta}(z))$. Here $f$ and $g$ have different set of network parameters, but for convenience we use $\theta$ to represent a concatenation of both network, which also indicates the two networks could be jointly trained. Ideally, the intermediate feature $g_{\theta}(z)$ simulates the real scene $I_t$ and $f_{\theta}$ simulate the physical interference such as blur and noise to approximate the observed image $I_s$, though $I_t$ may not directly exist in any layer of $f$. The first initiative is the following objective with a clear definition of the convolution kernel $\mathcal K$:
\begin{align}
    I_t,\theta^{\star}= \arg\min_{y,\theta} \| y \ast  \mathcal K -I_s \| +\beta \| y - g_{\theta}(z) \|.
\end{align}
We wish to reserve the highly strong expressivity of $g$ to simulate a complex enough scene and simultaneously make $f$ towards a simple function to mimic a physical corruption process. To further make the physical interference more complex, we substitute the convolution by $f$ and obtain: 
\begin{align}
    I_t,\theta^{\star}= \arg\min_{y,\theta} \|  f_{\theta}(y) -I_s \| +\beta \| y - g_{\theta}(z) \|.
\end{align}

\begin{figure*}[htbp]
	\centering
	\subfigure 
	{\includegraphics[width=0.9\textwidth]{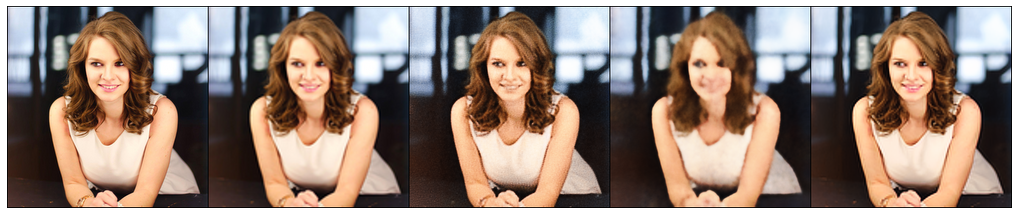}}
	\subfigure 
{\includegraphics[width=0.9\textwidth]{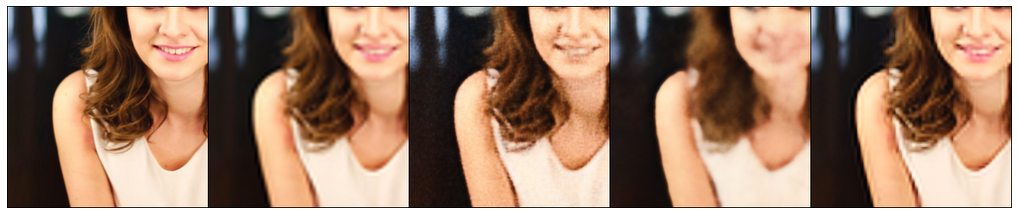}}
	\caption{ Visualization of Super-resolution with limited running time.  Left to right: high resolution, low resolution,   DIP (353 sec), DeepRED + TV (112.1 sec), DNCF (12.0 sec). 
	}
	\label{fig:SR2}
	\vspace{-5pt}
\end{figure*}
Here we pursue a smart initialization of $g_{\theta}(z)$,  
such as $y= G_{\rho} ( I_s) +\epsilon$, where $G$ could be Gaussian smoothing filters of standard deviation $\rho$ or bicubic interpolation, for specific tasks. This is to suggest that in the degrade case of $f$ being an identity mapping, the gradient of the objective function are nonzero. By initializing the intermediate feature at $y_0=   G_{\rho}  (I_s)$,  to approximate a feasible solution of image, but not too close to $I_s$ as to having a degenerated solution like $I_t=I_s$ and $f$ is an identity mapping. We therefore add additional regularization $\mathcal R$ and further refine the results by gradient descent, by jointly optimizing the following:
\begin{align*}
    \mathcal L(\theta,y)= \|  f_{\theta}(y) -I_s \|^2 +\beta \| y -  G_{\rho}  ( I_s) \|^2 + \mathcal R(\theta,y).
\end{align*}
We refer to our methods as the Deep Nonparametric Nonnegative Nonlinear Filters (DNCF). As it degenerates to a simple filter $G$ as a trivial solution, and if $\beta$ is small enough, it could lead to a fixed point solution as $y = f_{\theta}(y) =I_s$. We generate the real scene by choosing a better result (OPT) according to the photometric measurement, via heuristic selection. This prevents the hard cases of DIP being degenerated, e.g. the current result $f_{\theta}(y)$ within restricted running time still resembles random signal (as $y$ is randomly initialized), and in this case we take passive prediction $y$, or try a aggressive prediction $f_{\theta}(f_{\theta}(y^{\star}))$ for luck,
\begin{align*}
   I_t =OPT(y^{\star},f_{\theta}(y^{\star}),f_{\theta}(f_{\theta}(y^{\star})),\quad y^{\star}=\arg \min_y \min_{\theta}\mathcal L(y,\theta).
\end{align*}

\begin{figure*}[htbp]
	\centering
		\subfigure
	{\includegraphics[width=0.45\textwidth]{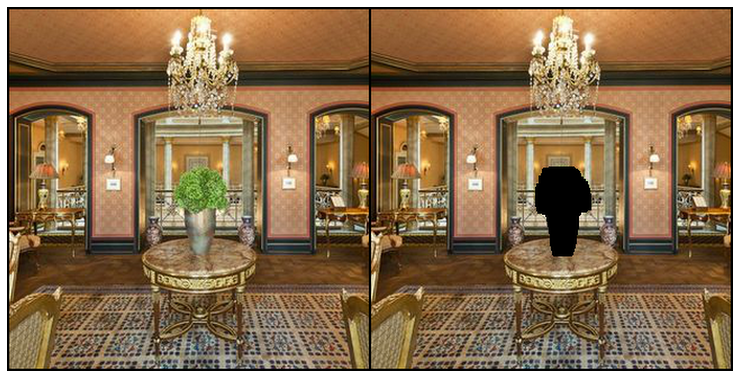}}
	\subfigure 
	{\includegraphics[width=0.225\textwidth]{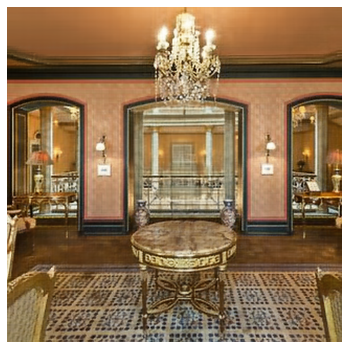}}
	\subfigure 
	{\includegraphics[width=0.225\textwidth]{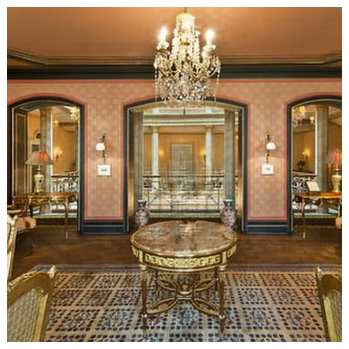}}
	\subfigure 
	{\includegraphics[width=0.45\textwidth]{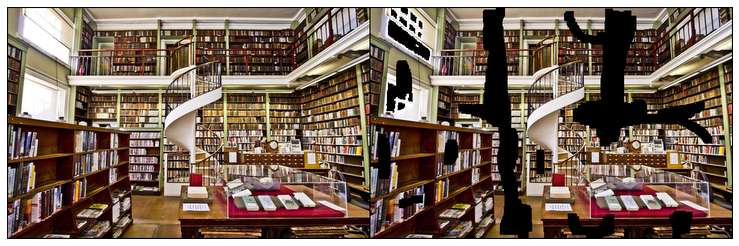}}
	\subfigure 
	{\includegraphics[width=0.225\textwidth]{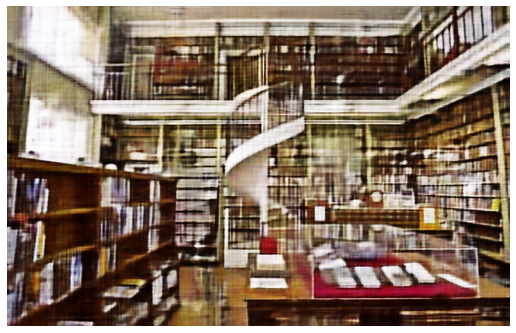}}
	\subfigure 
	{\includegraphics[width=0.225\textwidth]{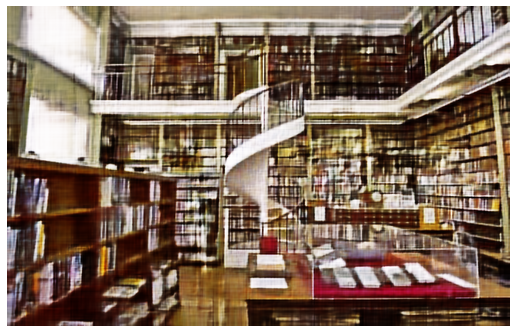}}
	\caption{ Visualization of image inpainting. From left to right: the original image, masked image, recovered images with DIP, and the results with D3NF. (\textit{inpaint1} on the top and \textit{inpaint2} on the bottom). 
	}
	\label{fig:inpaint}
	\vspace{-10pt}
\end{figure*}

Inspired by the previous work on regularization for blind deconvolution, we also seek a proper formulation of $\mathcal R$ ideally with convexity, and  easier inference. We note that the second quadratic term $\|y- g(I_s) \|^2$ contributes to the strong convexity to the objective function, and maintains consistency between two branches of prediction. We also include a idea that a non-negative combination of convex functions is still convex. By mathematical reduction, we know that a sequential network $f$ is convex w.r.t. the random vector $z$, given input data $I_s$, if all weights $\theta$ have non-negative elements, and the activation function like ReLU is convex and non-decreasing. The pioneering work from the input-convex network \cite{amos2017input} uses this rule and puts a strong nonnegative constraint on the parameters, which also limits the expressiveness of the network and therefore the final performance. 
In this paper, we use an interpolation between the convex network and an arbitrary one, through a soft regularization on the parameter weights. We use $\mathcal R(\theta)$ denotes the additional regularization to the negative weights 
\begin{align}\label{eq:reg}
\mathcal R(\theta) = \gamma \|\max(-\theta,0)\|.
\end{align}
The norm may change to $\ell_1$ or $\ell_2$. In an extreme case that $\gamma$ goes to infinity, then this becomes a constrained problem that requires all weights are nonnegative as \cite{amos2017input}.  Another middle way that goes between the rigorous input convexity and nonconvexity is to limit partial channels, or partial layers to have nonnegative weights: 
\begin{align}\label{eq:reg_part}
\mathcal R'(\theta) = \sum \gamma \|\max(-\theta[indx],0)\|,
\end{align}
where 
$indx$ is the indices of a subset of parameters. We typically only select those layers above a predefined depth to regularize, to keep lower layer being expressive. 

We use alternating optimization between the network parameters $\theta$ and the intermediate variables $y$, similiar to the Expectation-Maximization (EM). Although EM is nonconvex in general and is hard to optimize, a part of it might have an analytical solution. 
E.g., gradient descent can achieve a linear convergence rate on a strongly convex and smooth function; that is, to obtain a suboptimality gap decreases at an exponential speed\cite{nocedal2006numerical}. However, without the strong convexity, the decreasing rate may slow down inversely linear (for smooth and convex functions)\cite{nesterov1994interior}. Unfortunately, the (strong) convex property does not hold for the popular approaches based on neural networks. 

\section{Computational photography}
\textbf{Denoising.} 
Blind denoising is to recover the underlying clean image from a noisy observation without knowledge of the noise formation process. 
Assuming the additive noise model as $I_s = I_t + \epsilon$, where $I_s$ is source image for this task, i.e the noisy image and $I_t$ is the target image, i.e. clean image, and $\epsilon$ follows Gaussian distribution. DNCF optimizes the following objective function ($G$ as Gaussian smoothing):
\begin{align*}
  \mathcal L(\theta,y) = \|  \ft(y) -  I_s\|^2+ \beta\|y-G(I_s)\|^2+\mathcal R(\theta,y).
\end{align*}
\textbf{Texture synthesis and inpainting.} This technique \cite{portilla2000parametric}\cite{heeger1995pyramid}\cite{efros1999texture}\cite{efros2001image} is complementary to computational photography, if there is a defected region in pictures, and the synthesis and inpainting technique is hereafter to repair by filling in using textures from visually similar images. Some further applications include artistic manipulation of removing an object or individual from a picture while keeping the other pixels as photo-realistic as possible. There were several lines of classical approaches. 
For the inpainting problem with a picture $I_{s}$ that is removed pixels according to a mask $M \in \{0,1\}^{H \times W}$, DIP optimize the parameterization for the following objective function,
\begin{align*}
\mathcal L_{inpaint}(\theta,y) = \| M \odot \ft(y) - M \odot I_s\|^2+ \mathcal R(\theta,y)
\end{align*}
\textbf{Single image super-resolution (SR).} This is to generate higher resolution images from a lower resolution image restricted by camera sensors. If the resolution degradation problem is known to be from physical factors like blur effects, i.e. the observed image is a simple smoothed interpolation of the high-resolution target image, then SR algorithms could resemble image deblurring by alternating the estimation the blur kernel and deconvolution, such as blind image deconvolution as Eq.(\ref{eq:blur}).  
By denoting $I_s$ as the source image of lower resolution,  $D$ as the downsampling filter, $G$ as a one step SR filter, e.g. bicubic interpolation, DNCF-SR optimizes the following objective:
\begin{align*}
    \mathcal L_{SR}{(\theta,y)}=\| D(\ft(y)) -I_s\|^2+\beta\| y -G(I_s)\|^2+ \mathcal R(\theta,y).
\end{align*}
\textbf{Flash/no-flash photography.} 
Under some lousy lighting conditions, there will be a flash lighting control trade-off challenging to find. On the one hand, flash images contain unrealistic colors as photons from ambient illumination is comparably under-estimated, and this reduces the brightness of the natural color of objects, on the other hand, no-flash images have to rely on strong gains of cameras as photons are insufficient, and makes a stronger noise being inevitable as a byproduct. 
A representation algorithm \cite{petschnigg2004digital} is to use a joint bilateral filter on the no-flash image, where the flash image serves to provide edges information due to its advantage of less noise. To apply DIP on this task, we denote the flash image as $I_{f}$ and the no-flash image as $I_{nf}$, and let  $\epsilon$ follows multivariate Gaussian distribution and $\gamma$ to be the noise regularization, $G$ as the Gaussian smoothing filter, and DNCF-Flash optimizes the following:
\begin{align*}
\mathcal L_{flash}(\theta,y)=   \| \ft (y)- I_{nf}\|^2 +\beta\| y-G(I_f +  \epsilon)\|^2+\mathcal R(\theta,y).
\end{align*}
\begin{figure}[htbp]
	\subfigure 
		{\includegraphics[width=0.5\textwidth]{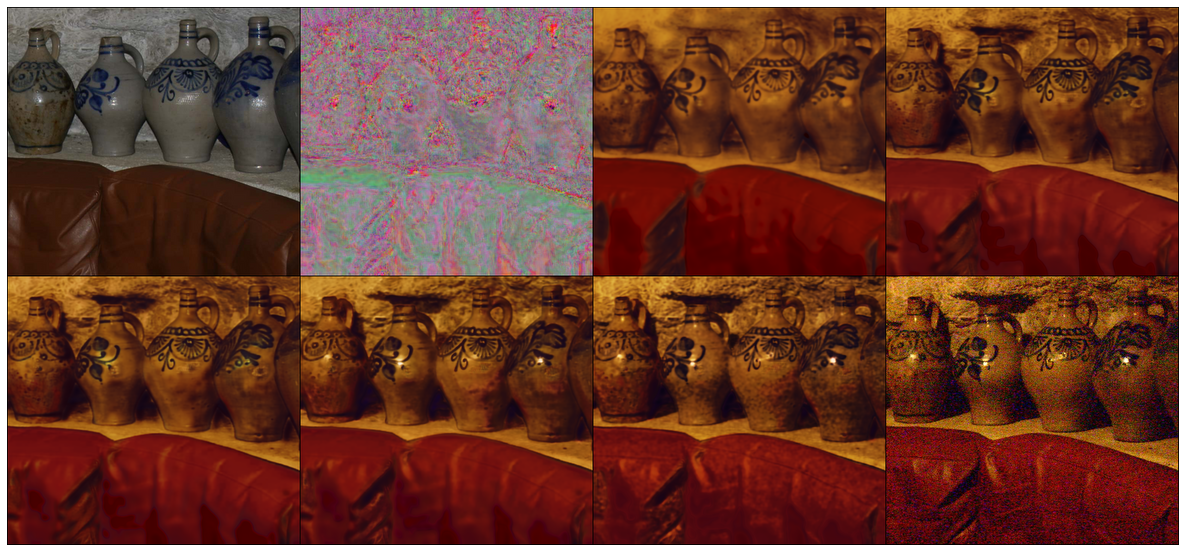}}
	\caption{ Visualization of the optimization procedure. In left-right, top-down order, flash image, initialized $I_t$, $5$ middle results by every 200 iterations, and then the original non-flash image. }
	\label{fig:flash_converge}
	\vspace{-10pt}
\end{figure}

\section{Adverarial defense for machine learning}
There are several ways to categorize adversarial attack methods. One perspective to classify is based on the final goal of the adversary. For example, poisoning attacks refer to add fake images and labels to make the overall trained models be poor performance \cite{biggio2012poisoning}, and this of course, requires access to the training sets and are plausible for web-crawled data. The evasion attacks give up on misleading the models but try to generate samples that are hard to recognize, and this resembles an extreme version of data augmentation, e.g. to put a mask on a traffic sign \cite{eykholt2018robust}. Some adversary attacks aim for fix classes of samples $I$, say with label $t$, and they try to generate visually similar samples $I'$ but with false label $t'$, and these are referred to as targeted attack, and the generalized attacks without specific label designation are non-targeted attacks. From another perspective, we could categorize the attacks by whether the adversary has access to the machine learning models configurations and parameters, whose leakage will boost the capability of attacks. White-box attacks refer to the methods with knowledge of the machine learning models. such as L-BFGS \cite{szegedy2013intriguing}, FGSM \cite{goodfellow2014explaining}, BIM \cite{kurakin2016adversarial} and CW \cite{carlini2017towards}.

\textbf{Fast gradient sign method (FGSM)} This was proposed in \cite{goodfellow2014explaining}. We denote the classifier model as $C$, and the misleading images $I^{adv}$ are generated from real samples $I$ and the true label $t$ as
\begin{align}
      I^{adv} =I+\epsilon \sign(\nabla_I \mathcal L(C(I),t)).
\end{align}
\textbf{Basic iterative method (BIM)} This \cite{kurakin2016adversarial} could be viewed as a multi-step variant of FGSM of smaller step size $\alpha$ with clipping to $\epsilon$-ball each iteration, which is a projecting operation to restrict the adversary examples being within $\epsilon$ distance from the original image $I_0$
\begin{align}
    I^{adv}_{n}=\textit{Proj}_{\epsilon}(I_{n-1}+\alpha\sign(\nabla_I\mathcal L(C(I_{n-1}), t))).
\end{align}
\textbf{Carlini $\&$ Wagner} The L-BFGS attacks was proposed in\cite{szegedy2013intriguing} which formulated the attack as an optimization algorithm for finding an optimal adversarial example $I$ that is predicted as an designated label $t$, 
which can be relaxed as an regularized optimization problem
\begin{align}
     \min_{I} \|I_t-I\| + \mathcal L(I,t),\quad s.t.\quad I \in [0,1]^m.
\end{align}
and solving this with L-BFGS. On top of the objective function,  Carlini and Wagner proposed the following margin loss in \cite{carlini2017towards}, to maximize the relative confidence of the label $t$ comparing to all other labels,
\begin{align}
     \min_{I} \|I_t-I\| +\max(\max_{t'} \mathcal L(I,t') -\mathcal L(I,t)  ,0),
\end{align}
\textbf{DNCF as adversary defense}
The  difference between the physical interference and adversary attacks is that the later case could be as bad as possible for classification. However, DNCF is insensitive to those pixel-wise changes, and will generate results randomly each time. DNCF optimizes the following function with $G$ as Gaussian smoothing and $\epsilon$ as random perturbation:
\begin{align*}
\mathcal L(\theta,y)=\|I^{adv}-f_{\theta}(y)\|^2+\beta \|G(I^{adv})+\epsilon-y\|^2,
\end{align*}
and perform the classification model on the denoising result $f_{\theta^{\star}} (y^{\star})$.  Comparing to computational photography, DNCF might be easier to remove adversary noises without the need to fixate on the pixel-wise accuracy for visual effects.

\begin{figure*}[htbp]\label{fig:cw_attack}
	\subfigure 
		{\includegraphics[width=1.\textwidth]{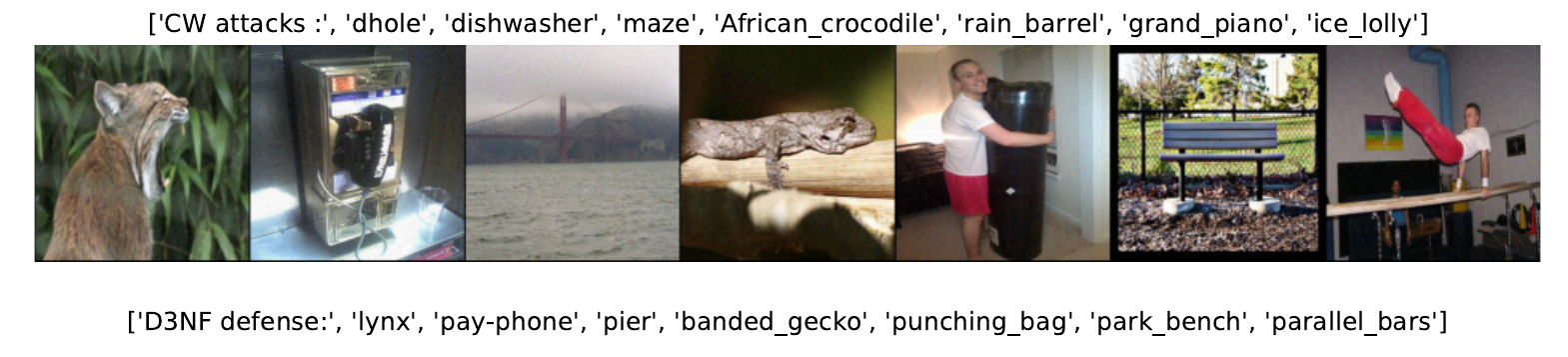}}
	\subfigure 
		{\includegraphics[width=1.\textwidth]{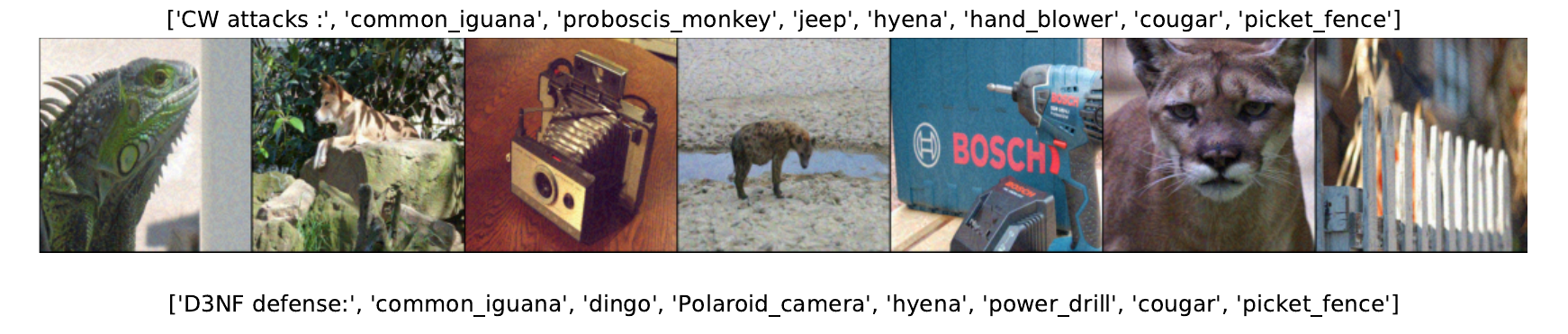}}
	\vspace{-10pt}
	\caption{  ImageNet data attacked by CW, with predictions of attacked images on the top and  defended predictions on the bottom. }
	\vspace{-10pt}
\end{figure*}
\section{Optimization perspectives}
This partial channel-wise convexification in Eq.(\ref{eq:reg_part}) aims to provide a weaker definition of strong convexity (SC), e.g., the constant nullspace strong convexity (CNSC) \cite{yen2014constant} \cite{zhong2014proximal}, and try to pursue a faster optimization algorithm under these conditions. CNSC rises naturally from the notation of the general convexity: as a function may not be strongly convex on the entire linear space $\mathcal Y$, it is still possible that the function is strongly convex on a subspace $\Phi$ of $\mathcal Y$; however, it may not be convex on the orthogonal space $\Phi^{\top}$. For notation, we use $\pro$ to represent the projection to a subspace $\Phi$. 
\begin{definition}
A function  $f(y)$ that is twice-differentiable w.r.t. $y \in \mathbb R^{K}$, is defined as having constant nullspace strong convexity of constant $m\geq 0$ if there exists a subspace $\Phi \in \mathbb R^{K'}$ on which the function $f(y)$ only depends, as $z = \pro(y)$ and $f'(z)=f(y)$, and the Hessian matrix $H(z)$ of $f'(z)=f(y)$ has the following property:
\begin{align}
v^{\top} H(z) v \geq m \|v\|^2, \quad
v^{\top}H(z) =0,\quad \forall v \in \Phi.
\end{align}
\end{definition}
We notice that if the subspace $\Phi$ is equivalent to $\mathbb R^{K}$, then CNSC equals to SC (Strong Convexity). Generally speaking, for the problem that takes much fewer data samples, technically, only one sample during inference, than the number of elements in that sample, CNSC is a more reasonable condition to satisfy since the inference procedure, since the gradient vector w.r.t. input lies in a much lower-dimensional subspace. 
For one last layer in the decomposition,
\begin{align}
\mathcal L(y) =\|\sigma(\theta_l^{\top} y +\theta^{\top}_l f_h(z)+b)-I_s\|^2_2 +\lambda \|y \|^2,
\end{align}
then the gradient vectors $\partial \mathcal L(y)$ lies in the subspace spanned by $\theta_l$, whose rank might very likely be much lower than $y$, which is typically high-dimensional by design.
\begin{lemma}\label{thm:subcvx}
If the objective function lies in subspace $\Phi$, that there exists an alternative expression of $L$, say $L'$ only depends on the subspace $\Phi$, such that $\mathcal L(y) = \mathcal L'(\pro(y))$, then the gradient vector $ p(y)$ and Hessian matrix $H$ also lies in the subspace $\Phi$.
\end{lemma}
We propose to use a quasi-Newton optimization algorithm to update the intermediate features $y$. The basic idea of this second-order algorithms is to minimize the second-order expansion around each iteration. Based on the the last iteration $y_t$, the gradient vector $p_t=\gt$, and an approximation $B_t \in \mathbb R^{D \times D}$ to the Hessian matrix, we give the ascent direction $d_t$ of this iteration by
\begin{align}
d_t= \arg\min_{d}p_t^{\top} d + \frac{1}{2} d^{\top} B_t d.
\end{align}
The step size for this iteration $\alpha_t$ is obtained by line search, which is iterative tried over $\{b^c\}, c \in [0,1,\cdots], b \in (0,1)$, from the largest to the smallest, to meet the Armijo rule of
\begin{align}
 \mathcal L(y^t+ \alpha_t d_t) \leq  \mathcal L(y^t) +\alpha_t \xi p_t^{\top} d_t.
\end{align}
where $\xi \in (0,1)$ is a constant. Applying BFGS algorithm \cite{dennis1977quasi} \cite{wright1999numerical}, the approximated Hessian $B_{t+1}$ for the next iteration is:
\begin{align}
B_{t+1} = B_{t} -\frac{B_{t} s_{t} s_{t}^{\top} B_{t}}{s_t^{\top} B_t s_t} +\frac{o_t o_t^{\top}}{o_t^{\top} s_t},
\end{align}
where $ s_t= y_{t+1}-y_t,\quad  o_t = p_{t+1} -p_t$. However, only updating $y$ leads to insufficient diversity in the result, as $y$ is constrained in the subspace $\Phi$, so we heuristically update $\theta$ and $y$ alternatively. 

\begin{table}[h]
\caption{Performance of DNCF for adversary defense: the running time (seconds) of attacks. the accuracy ($\%$) of the original network, the accuracy after being attacked, and the accuracy after defense. The running time of DNCF is 126 seconds for 200 images.}
\vspace{-5pt}
\label{table:defense}
\begin{center}
\begin{tabular}{c|c| c|  c| c c| c| c| c |c c c}
\hline
\hline
Attack   &time  &Ac(orig) &Ac(attack) &Ac(defense)\\
\hline
CW   & 1646.2  & 75.48  & 0.00  & 64.42\\
BIM   & 32.26  & 67.79  & 0.00  & 21.63\\
FGSM   & 5.60  & 72.60  & 18.27  & 33.65 \\
PGD  & 30.21  & 74.04  & 0.00  & 15.87 \\
FFGSM   & 5.45  & 75.00  & 16.83  & 30.29 \\
\hline
\hline
\end{tabular}
\vspace{-10pt}
\end{center}
\end{table}

\section{Experiments}
\begin{figure}[htbp]
	\subfigure 
		{\includegraphics[width=0.5\textwidth]{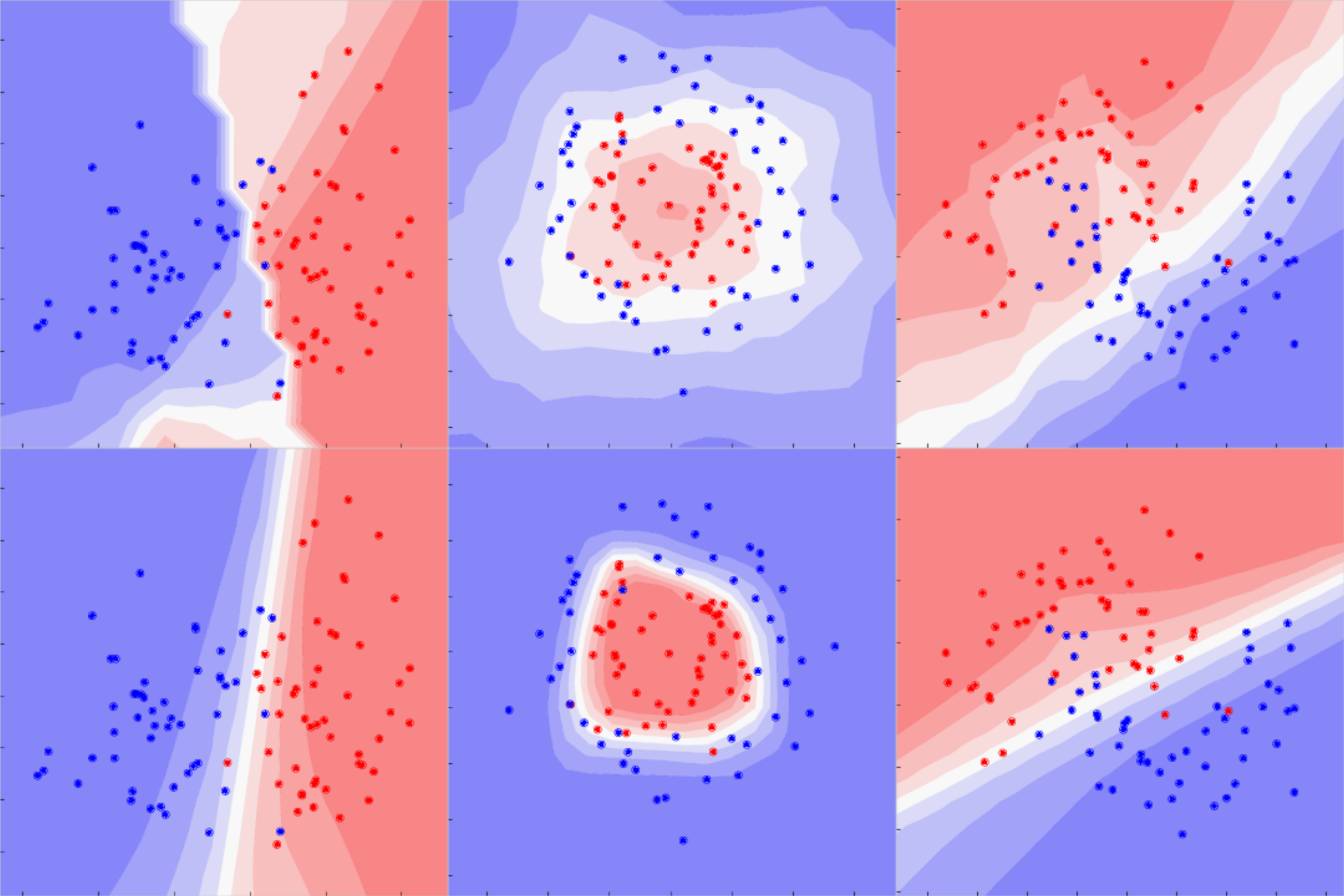}}
		\subfigure 
		{\includegraphics[width=0.5\textwidth]{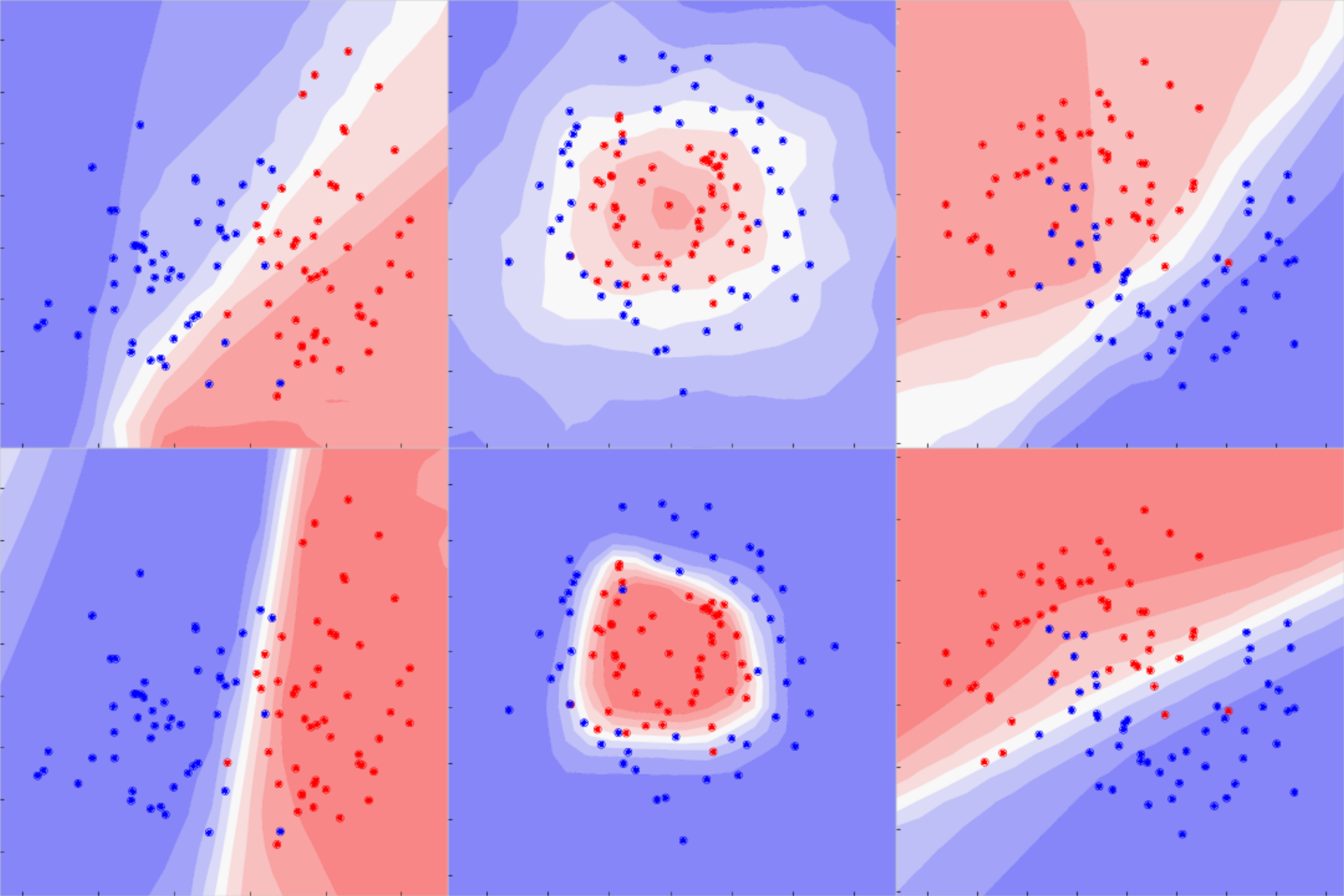}}
	\caption{ Visualization of prediction regions. 3 columns on the top: Nonnegative regularization;  3 columns on the bottom: ICNN.
	}
	\label{fig:icnn}
	\vspace{-10pt}
\end{figure}

\textbf{Visualization of convexification.} We first visualize the convexification effect of Eq.(\ref{eq:reg_part}) on the neural networks for smaller tasks. We give a simple example of synthetic data as the fully input convex network (FICNN) as described in \cite{amos2017input}, with a $2$-dimensional regression problem. This network takes both the data $z \in \mathbb R^{2}$ and a label proposal $y \in \{0,1\}$ as the network input and predicts the compatibility score of the proposal. The layers within the network have a special structure, in  the $l$th hidden layer, we require $z_l$ of $f$ has the following formulation,
\begin{align}
z_{l}=\sigma(\theta_l^z z_{l-1} + \theta_l^y y + b_{l}), \quad z_{l} \in \mathbb R^{K_l},
\end{align}
where $\sigma$ is ReLU function. The last layer projects to a scalar variable and the loss function is square distance. We name the network with our proposed regularization as CVXR-Net. FICNN uses the same network architecture with CVXR-Net, they both have two fully connected layers of $200$-dimensional latent features. Beside Partially ICNN (PICNN) is built on top of ICNN, except that latent features $z_l$ have a branch connecting $y$ that allows positive weights and therefore has relatively stronger expressive power. FICNN and PICNN use projected gradient descent for optimization to keep the network weights strictly non-negative. After training both networks to convergence, we demonstrate the decision boundary in figure(\ref{fig:icnn}) for several synthetic data sets, where red and blue points are from two different classes. We see that CVXR-Net produces a better boundary for harder examples and maintains a similar behavior on easier examples.
\begin{table}[h]
\caption{Performance comparison between DNCF (less than 700 iterations) and DIP (with unconstrained running time). Here $\downarrow$ means better images have smaller values and vice versa. }
\label{table:comp}
\begin{center}
\begin{tabular}{c|c| c|  c c c| c| c| c |c c c}
\hline
\hline
task   &DIP  &DNCF  &Figures\\
\hline
Denoise        &9.9e-4$\downarrow$ & 9.4e-3 $\downarrow$ &fig\ref{fig:denoise}\\
Flash   &6.887e-3 $\downarrow$  &6.705e-3 $\downarrow$ &fig\ref{fig:flash_converge}\\
Inpaint1 &7.08e-4$\downarrow$ &5e-4 $\downarrow$ &fig\ref{fig:inpaint}\\
Inpaint2   &9.521e-3$\downarrow$  & 8.4e-3$\downarrow$ &fig\ref{fig:inpaint} \\
SR   &19.067$\uparrow$   &19.447 $\uparrow$ &fig\ref{fig:SR}\\
\hline
\hline
\end{tabular}
\vspace{-10pt}
\end{center}
\end{table}

\textbf{Computational photography.} We conduct the experiments of DNCF with simpler version of the original implementation of DIP by smaller network. The visual results are in figures (\ref{fig:denoise} \ref{fig:SR} \ref{fig:inpaint}). We use the \textit{skip} network for all experiments, an encoder-decoder architecture with skip connections. The network consists of several blocks, and each block has an equal number of channels in each of the convolution layers. Each block has four convolution layers, four batch normalization layers, and four ReLU layers, where the second convolution layer has a stride of $\times 2$, therefore, it reduces the resolution and an upsampling layer follows to expand the resolution by $\times 2$. The last block also consists of an extra convolution layer to project the last feature tensor to $3$-channel that matches the images. For general applications, the output of the network is in the same resolution with the input, except that for super-resolution, there is a $\times 8$ resolution expansion effect, since we set to generate images in this resolution.  
We compare against total variation (TV) denoising \cite{chambolle2004algorithm}, bicubic interpolation, and DeepRED \cite{Mataev_2019_ICCV} combined to We restricted the algorithm to run for at most 400 iterations on NVIDIA GTX 1060 with 6GB graphics memory, as the original parameter of DIP runs considerably slow, as long as 300 seconds for a best result of super-resolution. We put the details of results for different tasks in Table.(\ref{table:comp}).
We start with $\gamma$ in the table as the regularization coefficient. Each time we found that the convexification regularization is larger than the mean square error (MSE) loss in forward-propagation, we reduce $\gamma$ to $1/4$ of its last value, to prevent that the regularization dominates the total loss. This adaptive regularization also saves us from extensive tuning $\gamma$ for each task individually. Our essence is to make the DIP practical on personal computers, therefore despite that, we inherit most hyperparameters, e.g, the number of layers, we reduce numbers of channels for those applications to fit in $4GB$ GPU memory, and reduce the number of extra iterations without visible changes. We report the performance in Table.(\ref{table:comp}). In the table, we show MSE of both DIP and DNCF; except for the super-resolution (SR) task, the metric is the peak SNR (PSNR) of the ground-truth against the SR version. Here  $\uparrow$ means the higher the better and $\downarrow$ means reversely. 
On the flash task,  we run $800$ iterations of optimizing the whole network and then only optimizing the last block and its $32$-channel intermediate features for about $1,200$ iterations using LBFGS, then compare with DIP baseline which is to optimize the whole network using Adam optimization, for the same period of time. We plot the convergence of MSE, the transition of the image flash-no flash photography in Figure(\ref{fig:flash_converge}), i.e. from no-flash to flash version, for every $200$ iterations.  Note that for the $5$ middle results of $I_t$, 
and $600$ iterations on a $32$-channel DNCF give a good enough result, in comparison to original DIP of $96$-channels with the same amount of iterations. The convexification improves the visual details and decreases MSE. In Table.(\ref{table:comp}) DNCF reduces MSE on all applications. We also see a great amount of information from the figures, e.g. in Figure.(\ref{fig:denoise},\ref{fig:SR}), DNCF generates images with weaker blur degradation and clearer quality; in Figure.(\ref{fig:inpaint}) DNCF fills in the hole with better patterns. 

\textbf{Adversary Defense.} We tested the effect of removing adversary attacks from images and use GoogleNet Inception V3 as the classifier. We randomly select 100 classes from the validation set of ImageNet ILSVRC 2012. We randomly tested 200 images for each methods of attacking (so the accuracy on clean images varies slightly), including the aforementioned methods, and projected gradient descent (PGD) \cite{madry2017towards} and Fast FGSM (FFGSM) \cite{wong2020fast}. We set the $\epsilon$ value (the norm of maximum pertubation on the images), $\alpha$ value (the step size) and $n$ vlaue (number of steps) as follows:  $\epsilon=8/256,\quad \alpha=2/256, \quad n=7$. We use a simple DNCF of only 3 convolution layers. In this case, we could use a batch of image and run them in parallel, so they actually share a network parameterization as we have lower standard in photometric measurement. 
We put the classification results on original images, attacked images and defended images on Table.(\ref{table:defense}). We show the running time comparison between attacking and defense as well. We notice that DNCF is extremely effective against CW, and have different improvement against others.
\section{Conclusion}
In this paper we propose a deep nonparametric filter (DNCF) as a general framework of modelling physical interference and adversary attacks on images, and lead to a solution of computational photography and adversary defense. 

{\small
\bibliographystyle{ieee_fullname}
\bibliography{iccv.bbl}
}

\end{document}